\documentclass[letterpaper, 10 pt, conference]{ieeeconf}  % Comment this line out if you need a4paper

\IEEEoverridecommandlockouts                              % This command is only needed if 
\usepackage{caption}
\usepackage{graphicx}
\usepackage{booktabs}
\usepackage{threeparttable}
\usepackage{amsmath} % assumes amsmath package installed
\usepackage{amssymb}  % assumes amsmath package installed
\usepackage{orcidlink}
\usepackage{tikz}
\usepackage{longtable}
\usepackage[backend=biber,style=numeric,doi=false,isbn=false,url=true]{biblatex}
\newcommand{\refnum}[1]{%
  \tikz[baseline=(char.base)]{
    \node[
      shape=circle,
      fill=black,
      inner sep=0pt,
      minimum size=10pt, % fixed circle size
      text=white,
      font=\footnotesize,
      align=center
    ] (char) {#1};
  }%
}

\title{\LARGE \bf
ARK-V1: An LLM-Agent for
Knowledge Graph Question Answering Requiring
Commonsense Reasoning*
}

\author{Jan-Felix Klein$^{1}$ \orcidlink{0000-0002-3704-9567} and Lars Ohnemus$^{1}$
\orcidlink{0009-0009-0612-506X}
\thanks{*This work has been funded by the Deutsche Forschungsgemeinschaft (DFG, German Research Foundation) – SFB 1574 – 471687386.}%
\thanks{$^{1}$Jan-Felix Klein and Lars Ohnemus are with the Institute of Material Handling and Logistics (IFL) at the Karlsruhe Institute of Technology (KIT), 76131 Karlsruhe, Germany 
        {\tt\small jan-felix.klein@kit.edu}}%
}

\begin{document}

\maketitle
\thispagestyle{empty}
\pagestyle{empty}

%%%%%%%%%%%%%%%%%%%%%%%%%%%%%%%%%%%%%%%%%%%%%%%%%%%%%%%%%%%%%%%%%%%%%%%%%%%%%%%%
\begin{abstract}
% motivation
Large Language Models (LLMs) show strong reasoning abilities but rely on internalized knowledge that is often insufficient, outdated, or incorrect when trying to answer a question that requires specific domain knowledge. Knowledge Graphs (KGs) provide structured external knowledge, yet their complexity and multi-hop reasoning requirements make integration challenging. 
% contribution
We present ARK-V1, a simple KG-agent that iteratively explores graphs to answer natural language queries. 
% experiments
We evaluate several not fine-tuned state-of-the art LLMs as backbones for ARK-V1 on the \textit{CoLoTa} dataset, which requires both KG-based and commonsense reasoning over long-tail entities.
% results
ARK-V1 achieves substantially higher conditional accuracies than Chain-of-Thought baselines, and larger backbone models show a clear trend toward better coverage, correctness, and stability.

\end{abstract}

%%%%%%%%%%%%%%%%%%%%%%%%%%%%%%%%%%%%%%%%%%%%%%%%%%%%%%%%%%%%%%%%%%%%%%%%%%%%%%%%
\section{INTRODUCTION}
Large Language Models (LLMs) that are pre-trained on massive text corpora have demonstrated remarkable reasoning capabilities across a wide range of natural language tasks. They are able to answer questions in a zero-shot, closed-book manner by relying on their internalized knowledge. However, this knowledge is often insufficient, outdated, or even incorrect. As a result, LLMs frequently struggle with hallucinations, factual inaccuracies, and a lack of specialized domain knowledge.

Knowledge Graphs (KGs) offer a promising solution by serving as structured, external sources of factual knowledge. Yet, effectively integrating KGs into reasoning tasks remains challenging \cite{pan_unifying_2024}. KGs can be large and complex, containing numerous irrelevant relations, and many queries require multi-hop reasoning across entities. Furthermore, most LLMs are not fine-tuned to operate directly on KG structures, limiting their effectiveness in leveraging such resources.

These challenges are particularly evident in closed-domain applications, such as business settings, where knowledge graphs capture highly specialized and constantly evolving information. In such cases, a general-purpose LLM cannot reliably provide accurate or up-to-date knowledge unless it is explicitly connected to the KG. This highlights the importance of developing AI agents that can effectively explore and utilize KGs as external knowledge sources.

In this work, we present an \textbf{A}gent for \textbf{R}easoning on \textbf{K}nowledge Graphs ARK-V1\setcounter{footnote}{1}\footnote{\url{https://github.com/JaFeKl/ark_v1}}, a simple yet effective agent that iteratively explores a knowledge graph and formulates intermediate reasoning steps to answer natural language queries. We evaluate ARK-V1 on the \textit{CoLoTa} dataset, a newly created benchmark which requires multi-hop reasoning combined with commonsense reasoning. 

The remainder of this paper is structured as follows. Section \ref{sec:related_work} reviews related work. Section \ref{sec:architecture} introduces the agent architecture of ARK-V1. Section \ref{sec:experiments} motivates and describes our evaluation experiments, followed by a discussion of results in Section \ref{sec:results}. Finally, Section \ref{sec:limitations} outlines limitations and directions for future work.

\section{Related Work}
\label{sec:related_work}
LLMs have recently been integrated into Knowledge Graph Question Answering (KGQA), with methods broadly categorized into semantic parsing (SP) and Information Retrieval (IR) methods \cite{ding_enhancing_2025}. 
SP methods translate natural language (NL) queries into formal logical expressions (e.g., SPARQL), whereas IR methods aim to retrieve and utilize relevant KG information in natural language form.

Among the IR-based methods, Li et. al \cite{li_graph_2023} address multiple-choice QA by first linearizing KG triples into natural language passages using templates, then applying hybrid retrievers to retrieve relevant passages. After reranking with a pre-trained cross-encoder, the top passages are fed into a fine-tuned language model to score answer choices.
Guo et al. \cite{guo_knowledgenavigator_2024} propose \textit{KnowledgeNavigator}, a three-stage framework combining hop-number estimation, iterative triple retrieval, and LLM-based answer generation. 
Similarly, Baek et al. \cite{baek_knowledge-augmented_2023} study single-shot prompting where the prompt includes pre-retrieved KG triples, evaluating how the quality of the provided triples influence the result.

A complementary line of research explores agent-based approaches for KG reasoning. Jiang et al. \cite{jiang_kg-agent_2024} introduce a KG agent that employs a toolbox of extraction, logic, and semantic tools, with a fine-tuned LLaMA7-7B model as backbone. Luo et al. \cite{luo_reasoning_2023} propose RoG, which uses an instruction-fine-tuned LLaMA2-Chat-7B to plan relation paths before retrieval and reasoning. Sun et al. \cite{sun_think--graph_2023} present Think-on-Graph (ToG), where an LLM agent iteratively generates reasoning paths that are aggregated into answers.

While these methods achieve strong results, they are typically evaluated on KGQA datasets where the graph encodes all the information needed to answer and where entities are already familiar to the LLM. By contrast, our work focuses on the \textit{CoLoTa} dataset, which requires combining KG-based reasoning with commonsense reasoning over long-tail entities. This setting highlights different strengths and weaknesses of LLM-based agents.

\section{AGENT ARCHITECTURE}
\label{sec:architecture}
\begin{figure}[t]
      \centering
%       \framebox{\parbox{3in}{We suggest that you use a text box to insert a graphic (which is ideally a 300 dpi TIFF or EPS file, with all fonts embedded) because, in an document, this method is somewhat more stable than directly inserting a picture.
% }}
      \includegraphics[scale=0.7]{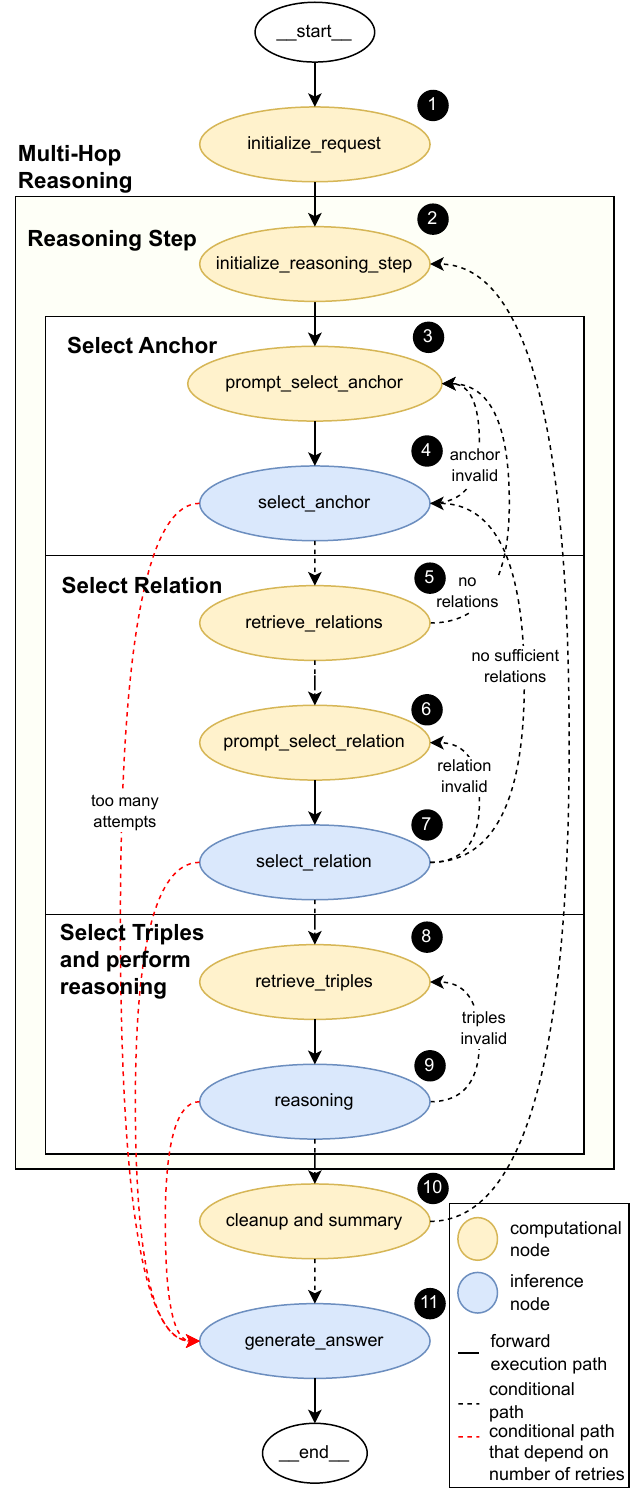}
      \caption{Agent Architecture of ARK-V1}
      \label{fig:agent_architecture}
\end{figure}
Given a natural language question $Q$, the goal of the ARK-V1 agent is to compute an answer $A$ by performing multi-hop reasoning over a knowledge graph (KG). The architecture, illustrated in Figure~\ref{fig:agent_architecture}, achieves this by iteratively constructing reasoning steps and summarizing them to produce the final answer.

We represent the KG as a \emph{property graph} \cite{hogan_knowledge_2021}, defined as
\begin{equation}
\mathcal{G} = \{(h, r, t, \phi) \mid h, t \in \mathcal{E}, \; r \in \mathcal{R}, \; \phi \in \Phi \},
\end{equation}
where $\mathcal{E}$ is the set of entities, $\mathcal{R}$ the set of relations, 
and $\Phi$ the space of edge properties. 
Each tuple $(h, r, t, \phi)$ represents a directed relation $r$ from head entity $h$ to tail entity $t$, 
optionally augmented with key--value attributes $\phi$ (e.g., temporal information, confidence scores, or provenance). 
\subsection{Initialization}
The agent is initialized (\refnum{1}) by adding the system prompt and the query $Q$ to the message context. 

\subsection{Reasoning Step}
At its core, the agent produces a sequence of reasoning steps $R^{(k)}$, where $k$ indexes the $k$-th step. Each $R^{(k)}$ is initialized in \refnum{2}, and its generation follows a fixed procedure outlined below.

\paragraph{Select Anchor}
The first sub-step is to select a single anchor entity from the knowledge graph $\mathcal{G}$ to begin exploration. An explanatory prompt is added to the message context (\refnum{3}), and the LLM proposes an anchor candidate $a^{(k,c)}$ (\refnum{4}), where $c$ denotes the $c$-th attempt to select an anchor within reasoning step $k$. 
An anchor candidate is considered valid if it corresponds to a head entity in the graph:
\begin{equation}
a^{(k,c)} \in \mathcal{E}_{\mathrm{head}} = \{ h \in \mathcal{E} \mid \exists (h, r, t, \phi) \in \mathcal{G} \} 
      \subseteq \mathcal{E}.
\end{equation}
The agent is routed according to the attempt counter $c$ and the validity of the selected anchor:
\begin{equation}
\text{route}(a^{(k,c)}) =
\begin{cases}
\refnum{11}, & c \geq C_{\max}, \\
\refnum{5}, & a^{(k,c)} \in \mathcal{E}_{\mathrm{head}}, \\
\refnum{3}, & \text{otherwise}.
\end{cases}
\end{equation}

\paragraph{Select Relation}
Conditioned on a valid anchor entity $a^{(k)}$, the set of outgoing relations $\mathcal{R}^{(k)}$ is retrieved in step \refnum{5}:
\begin{equation}
\mathcal{R}^{(k)} = \{ r \mid (h, r, t, \phi) \in \mathcal{G}, \; h = a^{(k)} \}.
\end{equation}
If no relations are found, the agent is routed back to the anchor selection step: 
\begin{equation}
\text{route}(\mathcal{R}^{(k)}) =
\begin{cases}
\refnum{3}, & \mathcal{R}^{(k)} = \emptyset, \\
\refnum{6}, & \text{otherwise}.
\end{cases}
\end{equation}

Next, a prompt describing the relation selection task is added to the message context (step \refnum{6}), and the LLM proposes a candidate relation $r^{(k,c)}$, where $c$ denotes the $c$-th attempt to select a valid relation. A candidate relation is considered valid if
\begin{equation}
    r^{(k, c)} \in \mathcal{R}^{(k)}.
\end{equation}
Routing proceeds based on the attempt counter and the validity of the selected relation:
\begin{equation}
\text{route}(r^{(k,c)}) =
\begin{cases}
\refnum{11}, & c \geq C_{\max}, \\
\refnum{4}, & r^{(k, c)} = \varnothing \\
\refnum{8}, &  r^{(k, c)} \in \mathcal{R}^{(k)}, \\
\refnum{6}, & \text{otherwise}.
\end{cases}
\end{equation}
\paragraph{Select Triples and Reasoning}
Next, the corresponding tuples ${T}^{(k)}$ given the selected anchor entity and relation  are retrieved in step \refnum{8}:
\begin{equation}
\mathcal{T}^{(k)} = \{ (h, r, t, \phi) \in \mathcal{G} \mid h = a^{(k)}, \; r = r^{(k)} \},
\end{equation}
Given the retrieved tuples, the agent is prompted to infer a reasoning step (\refnum{9}), defined as
\begin{equation}
R^{(k,c)} = \left( \mathcal{T}^{(k,c)}, \, i^{(k,c)}, \, f^{(k,c)} \right),
\end{equation}
where  $\mathcal{T}^{(k,c)} \subseteq \mathcal{T}^{(k)}$ is a selected set of tuples used for reasoning, $i^{(k,c)} \in \mathcal{L}$ is the inferred natural language implication on the reasoning step, and $f^{(k,c)} \in \{0,1\}$ is a boolean flag indicating whether to continue the reasoning process. Routing is performed as follows:
\begin{align}
\text{route}(R^{(k,c)}) =
\begin{cases}
\refnum{11}, & c \geq C_{\max}, \\
\refnum{10}, & \mathcal{T}^{(k,c)} \subseteq \mathcal{T}^{(k)} \\
\refnum{8}, & \text{otherwise}.
\end{cases}
\end{align}

\subsection{
Cleanup and Final Answer
}
After a valid reasoning step, a set of cleanup operations is performed (\refnum{10}):
\begin{enumerate}
    \item Generate a summary that includes all $k$ reasoning steps.
    \item Reset the message context to include only the system prompt, the query $Q$, and the generated summary, as initialized in step \refnum{1}.
\end{enumerate}
The agent is then routed according to the following conditions:
\begin{equation}
\text{route}(R^{(k)}) =
\begin{cases}
\refnum{11}, & k \geq K_{\max}, \\
\refnum{2}, & f^{(k,c)} = 1 \\
\refnum{11}, & \text{otherwise}.
\end{cases} \displaybreak
\end{equation}

Finally, the agent is prompted to produce the final answer $A$ based on the accumulated reasoning step summaries \refnum{11}.

\section{EXPERIMENTS}
\label{sec:experiments}
Existing LLMs demonstrate strong performance on entity-based commonsense reasoning benchmarks. This is largely because such benchmarks often include popular entities, for which abundant information is readily available online and thus incorporated into the models’ training data. However, to properly assess an agent’s ability to leverage a knowledge graph (KG) as a source of factual knowledge, the evaluation must rely on questions that are, ideally, unknown to the LLM. In contrast to the related methods, see Table \ref{tab:datasets}, we therefore evaluate ARK-V1 on the newly created \textit{CoLoTa} dataset \cite{toroghi_colota_2025}, which was explicitly designed with these characteristics in minds. 

\begin{table}[h]
\caption{Datasets used by LLMs+KGs methods}
\centering
\begin{tabular}{p{2cm} p{1cm} p{1.2cm} p{1cm}}
\toprule
\textbf{Methods} & \textbf{WebQSP} & \textbf{CWQ} & \textbf{GrailQA} \\
\midrule
RDPG \cite{ding_enhancing_2025} & X & X & - \\
RoG \cite{luo_reasoning_2023} & X & X & - \\
KG-Agent \cite{jiang_kg-agent_2024} & X & X & X \\
ToG \cite{sun_think--graph_2023} & X & X & X \\
\bottomrule
\end{tabular}
% \begin{tablenotes}
% \footnotesize
% \item $^{*}$ Baseline results from external sources.
% \end{tablenotes}
\label{tab:datasets}
\end{table}

\subsection{Dataset}
\begin{figure}[t]
      \centering
%       \framebox{\parbox{3in}{We suggest that you use a text box to insert a graphic (which is ideally a 300 dpi TIFF or EPS file, with all fonts embedded) because, in an document, this method is somewhat more stable than directly inserting a picture.
% }}
      \includegraphics[scale=0.7]{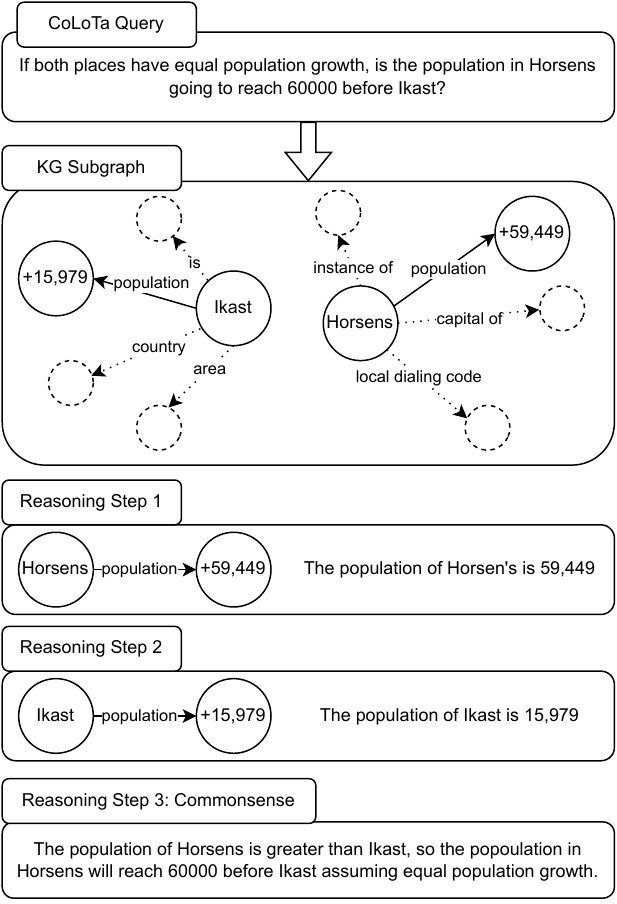}
      \caption{\textit{CoLoTa} \cite{toroghi_colota_2025} dataset entry example.}
      \label{fig:colota_example}
\end{figure}

\paragraph{CoLoTa}
We evaluate ARK-V1 on the CoLoTa dataset \cite{toroghi_colota_2025}, a recently introduced benchmark dataset designed to evaluate the entity-based commonsense reasoning abilities of LLMs in long-tail scenarios. CoLoTa contains 200 unique binary question-answer tasks with ground truth labels in $\{Truth, False\}$. The questions are adapted from StrategyQA \cite{geva_did_2021} and CREAK \cite{onoe_creak_2021}, but the original entities are replaced with long-tail entities to significantly increase reasoning difficulty. Successfully answering these queries requires leveraging a knowledge graph in combination with commonsense reasoning to augment the information explicitly present in the graph. Each QA instance provides not only the query and its ground-truth label, but also an inference rule, an ordered sequence of reasoning steps and a list of required reasoning skills (e.g., temporal reasoning, geography, sports) that are valuable for further analysis. Figure \ref{fig:colota_example} shows an example entry of the \textit{CoLoTa} dataset requiring two hops and a commonsense reasoning which requires number comparison.

\subsection{Metrics}
\paragraph{Answer Rate}
For the \textit{CoLoTa} dataset, the agent outputs one of three possible answers $\{\texttt{True}, \texttt{False}, \texttt{None}\}$. We define the \textit{answer rate} as the proportion of questions for which the agent provides a definite answer $\{\texttt{True} | \texttt{False}\}$. This metric captures the agent’s willingness to commit to an answer rather than abstaining with \texttt{None}.
\paragraph{Conditional Accuracy}
The fraction of correctly answered questions among those for which the agent provided a definite answer. This measures the accuracy of the agent when it chooses to answer. 
\paragraph{Overall Accuracy}: The fraction of correctly answered questions over the entire dataset, counting unanswered questions (\textit{None}) as incorrect. This reflects the trade-off between coverage and correctness.
\paragraph{Entropy-based Reliability}
\label{sec:reliability}
To quantify the stability of model predictions across repeated stochastic runs, we measure the distribution of answers $\{\texttt{True}, \texttt{False}, \texttt{None}\}$ for each question in CoLoTa. Let $p(a)$ denote the empirical probability of answer $a$ over 30 runs. We compute the Shannon entropy
\[
H = - \sum_{a} p(a)\,\log_2 p(a),
\]
which is maximized when the model’s outputs are evenly spread across all classes and minimized when the model consistently produces the same answer. To obtain a normalized reliability score between 0 and 1, we define
\[
\text{Reliability} = 1 - \frac{H}{\log_2 K},
\]
where $K$ is the number of possible answer classes (here, $K=3$). A score of 1 indicates perfectly consistent predictions across runs, while a score of 0 corresponds to maximal uncertainty (uniform predictions). We compute this score for each question and report the mean reliability across the entire dataset.

\section{RESULTS}
\label{sec:results}
The authors of \textit{CoLoTa} report baseline results using zero-shot and few-shot Chain-of-Thought (CoT) prompting across several models, with OpenAI's o1 model achieving the strongest performance. For reference, these baseline metrics are summarized in Table \ref{tab:baselines}. We observe that the high answer rates across models indicate a tendency to commit to an answer, even in cases where the underlying knowledge is not available. This suggests that while the models are unaware of the correct answers, they often default to guessing rather than abstaining, which inflates coverage but comes at the cost of accuracy.

\begin{table}[h]
\caption{Baseline results from the \textit{CoLoTa} authors \cite{toroghi_colota_2025}}
\label{table_baseline}
\centering
\begin{tabular}{p{2cm} p{1cm} p{1.7cm} p{1cm}}
\toprule
\textbf{Model} & \textbf{Answer Rate $\%$} & \textbf{Accuracy Conditional $\%$} & \textbf{Accuracy Overall $\%$} \\
\midrule
OpenAI-o1 (Zero-shot CoT)$^{*}$ & 97 & 65 & 67 \\
OpenAI-o1 (Few-shot CoT)$^{*}$ & 94 & 64 & 68 \\
\bottomrule
\end{tabular}
\begin{tablenotes}
\footnotesize
\item $^{*}$ Baseline results from external sources.
\end{tablenotes}
\label{tab:baselines}
\end{table}

Table \ref{table_example} summarizes the evaluation results on ARK-V1 on the \textit{CoLoTa} dataset under both stochastic ($T=0.7$, 30 runs) and deterministic conditions ($T=0$, single run). 
The stochastic experiment results of \textbf{Qwen3-8B}, \textbf{Qwen3-14B} and \textbf{Qwen3-30B} results show that larger models achieve higher conditional accuracy and overall stability. In particular, the entropy-based reliability increases from $0.52$ for \textbf{Qwen3-8B} to $0.65$ for \textbf{Qwen3-30B}, highlighting the improved consistency of larger models. It answers on average $\approx77\%$ of questions while achieving a conditional accuracy above $\approx91\%$, leading to an overall accuracy of $\approx70\%$.

For the deterministic results ($T=0$), we observe that Qwen3-235B, Gemini-2.5-Flash, GPT-5-Mini, and GPT-OSS-125B achieve accuracies that match or surpass the stochastic mid-scale models. Across these large backbones, conditional accuracy consistently exceeds $94\%$ with overall accuracies around $70$-$74\%$. These results indicate that scaling up backbone models yields diminishing gains in conditional accuracy but improves robustness in coverage and reliability. Interestingly, the mid-scale Qwen3-30B already approaches the performance of much larger models, suggesting that ARK-V1 can leverage KG structure effectively even without the very largest backbones.

\begin{table*}[t]
\caption{Evaluation results on the CoLoTa dataset \cite{toroghi_colota_2025}}
\label{table_example}
\centering
\begin{threeparttable}
\begin{tabular}{lccccccc}
\toprule
 & \multicolumn{4}{c}{\textbf{Temperature = 0.7; 30 runs}} & \multicolumn{3}{c}{\textbf{Temperature = 0; single run}} \\
\cmidrule(lr){2-5}\cmidrule(lr){6-8}
\textbf{Model} & \textbf{Answer Rate $\%$} & \multicolumn{2}{c}{\textbf{Accuracy $\%$}} & \textbf{Reliab.} & \textbf{Answer Rate $\%$} & \multicolumn{2}{c}{\textbf{Accuracy $\%$}} \\
\cmidrule(lr){3-4}\cmidrule(lr){7-8}
 &  & \textbf{Conditional} & \textbf{Overall} &  &  & \textbf{Conditional} & \textbf{Overall} \\
\midrule
% ----- Your LLMs: fill deterministic columns with your T=0 numbers -----
Qwen3-8B   & $57.67 \pm 2.68$ & $84.02 \pm 3.40$ & $48.42 \pm 2.49$ & 0.52&  59.00 & 82.20 & 48.50 \\
Qwen3-14B  & $45.93 \pm 4.81$ & $89.14 \pm 3.05$ & $40.95 \pm 4.53$ & 0.57 &  46.50 & 89.25 & 41.50\\
Qwen3-30B  & $76.98 \pm 2.31$ & $91.22 \pm 1.60$ & $70.20 \pm 1.67$ & 0.65 &  79.50 & 93.08 & 74.00 \\
\midrule
Qwen3-235B & %72.00             
& %97.22             
& %70.00             
&           &  71.50 & 95.80 & 68.50 \\
Qwen3 Next 80B A3B Instruct
& & & & &73.00 & 91.78& 67.00\\
Google Gemini 2.5 Flash & %77.50 
& %92.26 
& %71.50
& &76.50  &96.08  & 73.50 \\
OpenAI GPT-5 Mini & %77.50 
& %92.26 
& %71.50
&  &77.00  &94.81  & 73.00 \\
OpenAI oss 125B &  
&
&
&  &75.00  &94.00  & 70.50 \\
\bottomrule
\end{tabular}
\begin{tablenotes}
\footnotesize
\item Stochastic results show the mean value and the standard deviation over 30 runs; Reliability is the entropy-based stability score (higher is more stable). 
\end{tablenotes}
\end{threeparttable}
\end{table*}

\subsection{Error Analysis}
% Exemplary questions
Beyond simple failures of ARK-V1 to explore the graph or provide the correct answer, we identified two recurring sources of errors that reflect deeper challenges: (i) ambiguity in the formulation of questions, and (ii) conflicting evidence within the knowledge graph itself. Both challenges are specific to the CoLoTa dataset and highlight aspects that should be carefully addressed in future revisions.
\subsubsection{Ambiguity of questions}
During our analysis, we observed that some questions in CoLoTa are inherently ambiguous, which makes them challenging to answer. A representative example is
\textit{Entry S4: "Would it have been possible for Maria de Ventadorn to speak to someone 100 miles away?"}. The expected answer given the dataset is \texttt{False}, since in the 12th century no technology (e.g., telephones) existed that would have enabled direct communication over such a distance. However, different models used as a backbone of ARK-V1 interpreted the notion of “speaking” in divergent ways:
\begin{itemize}
    \item \textbf{GPT-5-Mini} returned \texttt{None}, arguing that the KG contains no explicit information about long-distance communication and therefore refrained from committing to an answer.
    \item \textbf{GPT-oss-125B} and \textbf{Qwen3-80B} reasoned that communication via messengers could be considered “speaking,” and thus returned \texttt{True}.
    \item \textbf{Qwen3-30B} interpreted the question as referring to mutual intelligibility of language, concluding that Maria de Ventadorn could have found someone 100 miles away who spoke the same language.
    \item The remaining models assumed “speaking” referred to direct, real-time conversation enabled by technology and therefore returned \texttt{False}.
\end{itemize}
Additional cases for this error include e.g. dataset entry \textit{S28} and \textit{S75}.
% Similar examples: 
% S28: "Can Valli di Comacchio fit in Gilan province?"
% S75: "Would you be likely to see storks at the baby shower of Hermann Hummels’ son in 1988?"

\subsubsection{Conflicting evidence in KG}
Another type of error arises when the knowledge graph itself contains conflicting or overlapping triples, leading to multiple valid reasoning paths with different conclusions. One examples is \textit{Entry 34: “Are any of Mahmoud Dowlatabadi’s works in the genre of The Makioka Sisters?”}
The expected reasoning sequence is:
\begin{itemize}
    \item Mahmoud Dowlatabadi $\rightarrow$ \textit{notable work} $\rightarrow$ Kelidar $\rightarrow$ \textit{genre} $\rightarrow$ novel
    \item The Makioka Sisters $\rightarrow$ \textit{genre} $\rightarrow$ novel
\end{itemize}
This path supports the correct answer \texttt{True}. However, the KG also contains the triple
\begin{itemize}
    \item Mahmoud Dowlatabadi $\rightarrow$ \textit{genre} $\rightarrow$ short story
\end{itemize}
which introduces ambiguity. Difference models handled this conflict in different ways:
\begin{itemize}
    \item \textbf{GPT-5-Mini}, \textbf{Qwen3-80B}, \textbf{Qwen3-30B} and \textbf{Qwen3-8B} stopped after exploring the \textit{short story} triple and concluding \texttt{False}.
    \item \textbf{GPT-oss-125B}, \textbf{Qwen3-14B} and \textbf{Qwen3-235B} initially considered the \textit{short story} relation but continued exploring, identifying Kelidar as a novel in the following and answering \texttt{True}.
    \item
    \textbf{Gemini-2.5-Flash} directly retrieved \textit{Kelidar} and answered \texttt{True} without being misled by the conflicting genre information.
\end{itemize}
% Similar examples:

% KG holds conflicting triples
% S34: "Are any of Mahmoud Dowlatabadi's works in the genre of The Makioka Sisters?"

% dataset reasoning sequence:

% Mahmoud Dowlatabadi->notable work->Kelidar->genre->novel
% Makioka Sisters->genre->novel

% however:
% Mahmoud Dowlatabad->genre->short story

% last path used by GPT-5 Mini, Qwen3-Next-80B, QWen3-30B, QWen3-8B

% \textbf{Gpt-oss-125B}, {Qwen3-14B}\textbf{Qwen3-235B} first explores short story but keeps exploring and finds Kelidar as a novel.

% \textbf{Gemini-2.5-Flash} directly looks for Kelidar

\subsubsection{Balancing Commonsense and KG Evidence}
A third type of error arises when the correct answer requires commonsense knowledge rather than explicit triples in the KG. For example, in \textit{Entry S193: “Is the name of the city where Francesco Renzi was born also a common human name?”}, most LLM backbones fail because the relevant subgraph contains no information about “Florence” being used as a personal name. In these cases, the agent over-relies on KG evidence and fails to draw on widely available commonsense knowledge. Additional cases for this type of error are e.g. dataset entries \textit{S119}, \textit{S163}, and \textit{S193}.

\section{LIMITATIONS AND FUTURE WORK}
\label{sec:limitations}
In this work, we introduced ARK-V1, an agent that iteratively explores a KG to answer natural language queries. On the \textit{CoLoTa} dataset, ARK-V1 outperformed Chain-of-Thought prompting baselines in both conditional and overall accuracy, with larger LLM backbones providing greater stability. Our analysis further revealed systematic errors linked to ambiguous questions, conflicting triples, and the challenge of balancing commonsense reasoning with KG evidence.

As a first version, ARK-V1 faces several limitations: token usage grows with exploration depth, redundant traversal of triples may occur, and prompting strategies remain relatively simple. Future work will focus on efficiency improvements, adaptive prompting, and applications to domain-specific graphs such as scene graphs in robotics or enterprise KGs.

\addtolength{\textheight}{-12cm}   % This command serves to balance the column lengths

%%%%%%%%%%%%%%%%%%%%%%%%%%%%%%%%%%%%%%%%%%%%%%%%%%%%%%%%%%%%%%%%%%%%%%%%%%%%%%%%

%%%%%%%%%%%%%%%%%%%%%%%%%%%%%%%%%%%%%%%%%%%%%%%%%%%%%%%%%%%%%%%%%%%%%%%%%%%%%%%%



% \section*{ACKNOWLEDGMENT}

% The preferred spelling of the word ÒacknowledgmentÓ in America is without an ÒeÓ after the ÒgÓ. Avoid the stilted expression, ÒOne of us (R. B. G.) thanks . . .Ó  Instead, try ÒR. B. G. thanksÓ. Put sponsor acknowledgments in the unnumbered footnote on the first page.

% %%%%%%%%%%%%%%%%%%%%%%%%%%%%%%%%%%%%%%%%%%%%%%%%%%%%%%%%%%%%%%%%%%%%%%%%%%%%%%%%

% References are important to the reader; therefore, each citation must be complete and correct. If at all possible, references should be commonly available publications.

% \bibliography{bibliography.bib}
\printbibliography

\end{document}